\documentclass[letterpaper, 10pt, conference]{ieeeconf}

\IEEEoverridecommandlockouts
\usepackage{cmap}           
\usepackage[utf8]{inputenc}
\usepackage[T1]{fontenc}
\usepackage{amsmath,amssymb,amsfonts}
\usepackage{graphicx}
\usepackage{booktabs}
\usepackage{multirow}
\usepackage{algorithm}
\usepackage{algorithmic}
\usepackage{color}
\usepackage{hyperref}
\usepackage{cite}
\usepackage{bm}
\usepackage{subcaption}
\usepackage{xspace}
\usepackage{tabularx}

\newcommand{\ours}{SoftNav\xspace}
\newcommand{\eg}{\emph{e.g.}\xspace}

\title{\LARGE \bf SoftNav: Injecting 3D Scene Tokens into VLMs\\for Embodied Navigation}

\author{
  Yi Wu$^{1}$, Junjie An$^{1}$, Xiao Liu$^{1}$, Yiqun Zhou$^{1}$, Yuechen Wu$^{2}$, \\
  Xiaoqing Guan$^{1}$, Shuyang Yu$^{1}$, You Wang$^{1*}$, and Guang Li$^{1}$\\
  $^{1}$Zhejiang University, $^{2}$Shandong University%
  \thanks{$^{*}$Corresponding author: You Wang, king\_wy@zju.edu.cn}%
}

\begin{document}

\maketitle
\thispagestyle{empty}
\pagestyle{empty}

\begin{abstract}
In goal-directed embodied navigation, where an agent must locate a specified target in an unseen environment, 3D scene understanding and navigation reasoning must work in concert.
Current approaches transmit 3D scene information to vision-language models (VLMs) through text, suggesting a representation gap in our tested configurations; a controlled ablation confirms that direct embedding-level transfer significantly outperforms the evaluated text serialization formats.
We introduce \ours, which injects entity-level 3D continuous representations---one token per detected object or frontier---into a VLM's hidden space as soft tokens through a lightweight projector.
With the 3D encoder and VLM frozen, only ${\sim}1{,}200$ samples and ${\sim}17$M trainable parameters are needed.
On HM3D-OVON, \ours achieves 74.2\%/68.3\%/66.7\% SR across three splits, surpassing all prior methods in both SR and SPL; the same navigation policy transfers zero-shot to GOAT-Bench (67.2\% SR), SG3D (47.2\% s-SR), and real-world robot deployment without retraining or architectural modification.
Injecting 3D scene tokens directly into VLMs bridges the representation gap, enabling transferable navigation with minimal training.
\end{abstract}

\section{INTRODUCTION}
\label{sec:intro}

Goal-directed embodied navigation requires an agent to locate specified targets in unseen 3D environments without prior maps.
Recent benchmarks have expanded goal specifications from fixed object vocabularies to open-vocabulary categories (OVON~\cite{yokoyama2024ovon}), multi-modal goals (GOAT-Bench~\cite{khanna2024goat}), and scene-graph-grounded targets (SG3D~\cite{sg3d}), placing increasing demands on 3D understanding and decision-making.
How this 3D information reaches the decision module---and what is lost in transit---is a critical yet underexplored question.

\begin{figure}[!t]
  \centering
  \includegraphics[width=\columnwidth]{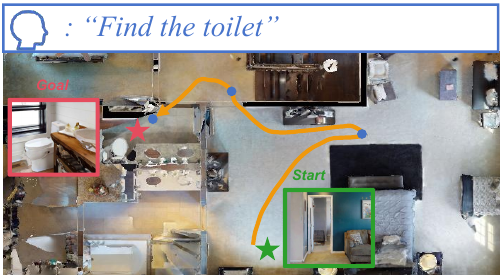}
  \caption{Given ``Find the toilet,'' the \ours agent navigates by selecting frontier waypoints guided by soft tokens injected from a frozen 3D scene encoder.}
  \label{fig:teaser}
\end{figure}

Two complementary paradigms have driven progress: modular approaches~\cite{chaplot2020semexp,yokoyama2024vlfm} decompose navigation into independently optimized perception and planning components.
MTU3D~\cite{mtu3d}, through its PQ3D encoder~\cite{zhu2024pq3d}, achieves state-of-the-art results across multiple benchmarks by encoding objects and frontiers into unified query embeddings.
In parallel, VLM-based methods~\cite{zhou2024navgpt,navid2024} exploit large-model reasoning for navigation but lack structured 3D scene perception.
Existing approaches that combine 3D perception with LLM reasoning transmit spatial information as text---serialized coordinates, scene-graph strings, or natural-language summaries~\cite{l3mvn,saynav,sg_nav,zhou2024navgpt}---but the information cost of this conversion has not been quantified.

To quantify this cost, a controlled ablation holds the 3D encoder and VLM constant while replacing the text interface with direct soft-token injection.
The result reveals a statistically significant performance penalty from the tested text serialization formats, particularly in path efficiency (\S\ref{sec:ablation}). This indicates a \emph{representation gap} between the 3D encoder's continuous features and the discrete text prompts, motivating a direct embedding-level alternative.

\ours realizes this transfer by projecting PQ3D's entity-level query embeddings---each encoding a semantically distinct object or frontier---into the VLM's hidden space as soft tokens through a lightweight MLP projector.
Each soft token fuses geometric, semantic, and spatial-relational information from cross-modal 3D grounding---content that text serialization cannot preserve.
These tokens are scattered into the VLM's input sequence alongside egocentric visual memory and a textual navigation prompt; the VLM autoregressively generates a target coordinate that is snapped to the nearest eligible frontier waypoint~\cite{yamauchi1997frontier}.
Training freezes both the 3D encoder and VLM; only the projector and LoRA adapters (${\sim}17$M parameters) are trained on ${\sim}1{,}200$ supervised samples distilled from an existing navigation system.

On HM3D-OVON, \ours achieves 74.2\%/68.3\%/66.7\% SR across three splits, surpassing all prior methods; the same navigation policy generalizes zero-shot to GOAT-Bench (67.2\% SR), SG3D (47.2\% s-SR), and real-world robot deployment without retraining or architectural modification.
These results, combined with the ablation findings, indicate that a primary representation gap in current pipelines lies in the text-based interface rather than in model capacity alone.

Our main contributions are:
\begin{itemize}
  \item We identify and quantify a \emph{representation gap} in VLM-based navigation: controlled ablation shows that text serialization incurs a statistically significant performance penalty---neither enriching the text channel nor model adaptation alone closes it.
  \item We introduce \ours, the first method to inject entity-level representations from a dedicated 3D scene encoder into a VLM's hidden space for navigation, via a lightweight MLP projector, requiring only ${\sim}17$M trainable parameters and ${\sim}1{,}200$ samples with both encoder and VLM frozen.
  \item A single \ours policy, trained only on OVON, achieves state-of-the-art SR and SPL (surpassing the prior best by 19--26\,pp SR), generalizes zero-shot to GOAT-Bench and SG3D, and transfers to real-world robot deployment---without retraining or architectural modification.
\end{itemize}

\begin{figure*}[t]
  \centering
  \includegraphics[width=\textwidth]{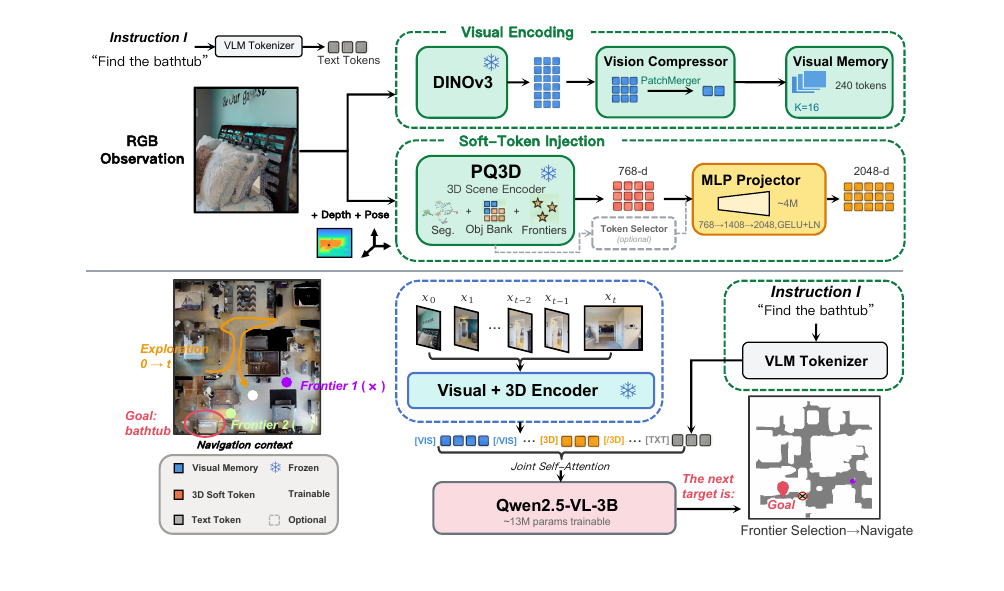}
  \caption{Overview of \ours. PQ3D extracts query embeddings from streaming RGB-D observations; the MLP projector maps them into the VLM's hidden space as soft tokens. The VLM jointly attends over visual memory, soft tokens, and text to select a frontier waypoint; only the projector and adapters are trained.}
  \label{fig:architecture}
\end{figure*}

\section{RELATED WORK}
\label{sec:related}

\textbf{3D Scene Information Transfer for Navigation.}
Goal-directed navigation tasks have been addressed by diverse architectures, including semantic-map planners~\cite{chaplot2020semexp,ramakrishnan2022poni}, VLM-scored frontier selection~\cite{yokoyama2024vlfm}, imitation--reinforcement hybrids~\cite{ramrakhya2023pirlnav}, video-based VLM policies~\cite{zhang2024uninavid}, and unified 3D perception~\cite{mtu3d}.
Yet a critical interface design choice has received limited scrutiny: \emph{how does 3D scene information reach the decision module?}
Existing answers fall into three categories.
SemExp~\cite{chaplot2020semexp} and PONI~\cite{ramakrishnan2022poni} convey spatial information through rasterized semantic maps consumed by learned policies.
A second category serializes structured 3D information into text: SayNav~\cite{saynav}, SG-Nav~\cite{sg_nav}, and VoroNav~\cite{wu2024voronav} convert scene graphs or topological structures into LLM-consumable strings, while NavGPT~\cite{zhou2024navgpt}, L3MVN~\cite{l3mvn}, MapGPT~\cite{chen2024mapgpt}, and ESC~\cite{zhou2023esc} render frontier coordinates, observations, or commonsense priors as natural-language prompts.
MTU3D~\cite{mtu3d} takes a different route: it feeds PQ3D's~\cite{zhu2024pq3d} 768-dimensional multi-modal query embeddings---fused from voxel geometry, multi-view appearance, and text---directly into a compact learned policy, bypassing text interfaces but also forgoing VLM reasoning.
These approaches either compress rich continuous features into text or restrict them to specialized policies, yet the performance cost of text serialization has not been empirically measured.
\ours provides the first controlled measurement and the first soft-token bridge between a dedicated 3D scene encoder and a VLM for navigation.

\textbf{Soft-Token Injection for Cross-Modal Alignment.}
Flamingo~\cite{alayrac2022flamingo} pioneered visual token injection into frozen LLMs via gated cross-attention. LLaVA~\cite{liu2023llava} and BLIP-2~\cite{li2023blip2} subsequently demonstrated that lightweight trainable projectors can align 2D visual encoder outputs with frozen LLM input spaces, establishing the soft-token injection paradigm.
In the 3D domain, 3D-LLM~\cite{hong20233dllm} and LEO~\cite{huang2024leo} project 3D point cloud features into LLM hidden states for scene understanding, and LL3DA~\cite{chen2024ll3da} extends visual instruction tuning to omni-3D tasks---but none injects structured 3D scene encoder outputs for online navigation decisions.
NaVid~\cite{navid2024} processes egocentric video streams directly through a VLM, and EmbodiedGPT~\cite{mu2023embodiedgpt} employs chain-of-thought reasoning for embodied planning---but both operate on 2D visual information without structured 3D scene representations.
Our \ours extends soft-token injection from 2D image patches to 3D entity-level representations. Unlike prior projectors that map fixed-grid visual patches encoding spatially local appearance, each of our soft tokens encodes a semantically distinct object or frontier produced by cross-modal 3D grounding.
The concurrent Dynam3D~\cite{wang2025dynam3d} injects hierarchical 3D tokens into a VLM for vision-and-language navigation, but constructs these tokens by projecting 2D visual features into 3D space rather than using a dedicated 3D scene encoder, and targets instruction-following rather than open-vocabulary object-goal navigation with frontier-based waypoint selection.
Table~\ref{tab:comparison} summarizes how existing paradigms compare along these design axes.

\begin{table}[t]
\centering
\caption{Comparison of navigation paradigms. \ours is the first to combine structured 3D perception with soft-token injection into a VLM. $\circ$\,=\,partial; *\,concurrent work.}
\label{tab:comparison}
\setlength{\tabcolsep}{2pt}
\small
\begin{tabular}{l ccccc}
\toprule
\multirow{2}{*}{Paradigm} & Vis. & Obj. & Front. & VLM & 3D$\to$VLM \\
       & Mem. & Per. & Expl. & Rea. & Interf. \\
\midrule
Sem.\ Map~\cite{chaplot2020semexp,ramakrishnan2022poni} & Map & \checkmark & \checkmark & -- & -- \\
3D Policy~\cite{mtu3d} & -- & \checkmark & \checkmark & -- & -- \\
Text+VLM~\cite{zhou2024navgpt,l3mvn,saynav} & Text & -- & $\circ$ & \checkmark & Text \\
VLM Frontier~\cite{yokoyama2024vlfm} & -- & -- & \checkmark & \checkmark & -- \\
Video VLM~\cite{navid2024,zhang2024uninavid} & Vid. & -- & -- & \checkmark & -- \\
3D Inject.*~\cite{wang2025dynam3d} & Vid. & $\circ$ & -- & \checkmark & Soft \\
\midrule
\textbf{\ours (Ours)} & \textbf{Img.} & \textbf{\checkmark} & \textbf{\checkmark} & \textbf{\checkmark} & \textbf{Soft} \\
\bottomrule
\end{tabular}
\end{table}

\section{METHOD}
\label{sec:method}

\subsection{System Overview}
\label{sec:overview}

Fig.~\ref{fig:architecture} illustrates the \ours pipeline: streaming RGB-D observations flow through two parallel pathways, a visual stream and a 3D scene stream, before converging with a textual navigation prompt in the VLM's hidden space.
The visual stream adopts the compression pipeline from~\cite{astranav}, which provides compact, training-free visual tokens: a frozen DINOv3~\cite{simeoni2025dinov3} backbone extracts patch features from each frame $I_t$, and a CompressionHead with PatchMerger reduces them to $N_v{=}15$ tokens:
\begin{equation}
  V_t = \text{PatchMerger}\!\bigl(\text{CompHead}\!\bigl(\text{DINOv3}(I_t)\bigr)\bigr) \;\in\; \mathbb{R}^{15 \times 2048}.
  \label{eq:vt}
\end{equation}
A rolling buffer of the most recent $K{=}16$ frames, set empirically, forms the visual memory $\mathcal{M}_t = [V_{t-K+1};\ldots;V_t] \in \mathbb{R}^{240 \times 2048}$.
The 3D stream extracts object and frontier query embeddings from accumulated RGB-D observations via a frozen scene encoder (PQ3D~\cite{zhu2024pq3d,mtu3d}), and a trainable MLP projector maps each 768-d embedding into the VLM's 2048-d hidden space as a soft token.
The VLM backbone (Qwen2.5-VL-3B~\cite{bai2025qwen2vl}) jointly attends over visual memory tokens, soft tokens, and a textual navigation prompt to select a frontier waypoint for execution.
Only the MLP projector (${\sim}4$M) and LoRA adapters~\cite{hu2022lora} (${\sim}13$M) are trainable; the visual encoder, 3D encoder, and VLM backbone remain frozen.

\subsection{PQ3D 3D Scene Encoder}
\label{sec:pq3d}

We adopt PQ3D~\cite{zhu2024pq3d}, a 3D vision-language model adapted for embodied navigation by MTU3D~\cite{mtu3d}.
From streaming RGB-D observations, PQ3D performs incremental 3D instance segmentation, maintains a persistent object bank, and detects exploration frontiers~\cite{yamauchi1997frontier} from the occupancy map.
Via cross-modal attention, PQ3D produces a 768-dimensional embedding for each detected object and frontier, yielding a unified set $\{q_i\}_{i=1}^{N} \in \mathbb{R}^{N \times 768}$, where $N = N_{\text{obj}} + N_{\text{frontier}}$.
Each 768-dimensional embedding encodes geometric, semantic, and spatial-relational information that two scalar coordinates cannot express.
PQ3D remains frozen throughout training; with only ${\sim}1{,}200$ samples, end-to-end fine-tuning risks destroying its pretrained representations.

\subsection{Soft-Token Injection}
\label{sec:soft_token}

\subsubsection{Projector Design}
The 768-d query space and the VLM's 2048-d hidden space can be aligned through cross-attention layers or simple linear/MLP projectors; we choose an MLP for two reasons.
First, our training set comprises only 1{,}187 samples---orders of magnitude smaller than the training data of either PQ3D or the VLM---so high-parameter cross-attention is prone to overfitting.
Second, PQ3D's query embeddings are already semantically rich (\S\ref{sec:pq3d}), so the remaining alignment gap is primarily geometric (dimensionality and distribution matching) rather than semantic---the same condition under which LLaVA~\cite{liu2023llava} showed a lightweight MLP suffices.

\subsubsection{MLP Projector}
We adopt a two-layer MLP with GELU activation and LayerNorm:
\begin{equation}
  \hat{q}_i = \text{LayerNorm}\!\bigl(W_2 \cdot \text{GELU}(W_1 q_i + b_1) + b_2\bigr),
  \label{eq:proj}
\end{equation}
where $W_1 \in \mathbb{R}^{1408 \times 768}$, $W_2 \in \mathbb{R}^{2048 \times 1408}$.
The intermediate dimension 1408 is the midpoint of $\text{linspace}(768, 2048, 3)$, a simple heuristic for gradual dimensionality expansion.

\subsubsection{Scatter Injection and Token Ordering}
During tokenization, special placeholder tokens are inserted into the text template at positions designated for 3D information; at the embedding level, these placeholders are replaced by the projected soft tokens, ensuring positional alignment with the VLM's learned position encodings.
The resulting input embedding sequence combines visual memory tokens $\mathcal{M}_t$, soft tokens, and text tokens $\mathcal{T}$:
\begin{equation}
  \mathcal{H}_{\text{input}} = \bigl[\mathcal{M}_t;\; \hat{q}_1, \ldots, \hat{q}_{N};\; \mathcal{T}\bigr].
  \label{eq:hinput}
\end{equation}
Within the soft-token block, frontier tokens precede object tokens as an ordering heuristic; in practice, truncation rarely occurs since all $N$ tokens typically fit within the token budget.
All $N$ tokens are injected by default; token selection strategies are examined in \S\ref{sec:ablation}.
Architecturally, the injection mechanism requires only fixed-dimensional embeddings and is not specific to PQ3D; validation with alternative encoders is left to future work (\S\ref{sec:conclusion}).

\subsection{Frontier Decision}
\label{sec:frontier}
Rather than treating the VLM's generated coordinate as a final navigation target, we snap it to the nearest eligible frontier waypoint~\cite{yamauchi1997frontier} detected from the occupancy map, ensuring the target is always reachable and lies on the boundary between explored and unexplored space.
Given the VLM-predicted coordinate $\hat{c}$, the selected frontier is:
\begin{equation}
  f^{*} = \underset{f_j \in \mathcal{F}_t}{\arg\min}\; \lVert \hat{c} - f_j \rVert_2, \quad \text{s.t.} \quad \frac{\lVert f^{*} - p_t \rVert_2}{\min_{f_k} \lVert f_k - p_t \rVert_2} \leq r,
\end{equation}
where $p_t$ is the agent's current pose and $r{=}2$ is a distance-ratio guardrail, set empirically, that prevents the agent from selecting an excessively distant frontier when closer candidates exist, mitigating compounding path-length errors.
When the frontier detector returns an empty set (\eg in fully explored rooms), a rescue fallback re-triggers exploration from the agent's current position.
A GreedyGeodesicFollower then executes local navigation to the selected frontier.

\subsection{Training Strategy}
\label{sec:training}

\subsubsection{Training Configuration}
We freeze the visual encoder, PQ3D, and the VLM backbone, training only the MLP projector and LoRA adapters applied to both attention and feed-forward layers.
Training uses ${\sim}1{,}200$ supervised samples collected from successful episodes of the MTU3D system~\cite{mtu3d} on OVON training scenes.

\subsubsection{Cross-Architecture Knowledge Transfer}
Training data are derived from frontier selections made by the PQ3D-based navigation system~\cite{mtu3d}.
Since frontier coordinates are architecture-independent, the teacher's selection at each decision step serves directly as the coordinate label for the VLM's autoregressive training objective:
\begin{equation}
  \mathcal{L} = -\sum_{t} \log P_{\theta}\!\left(y_t \;\middle|\; \mathcal{H}_{\text{input}}^{(t)},\; y_{<t}\right),
\end{equation}
where $\theta$ comprises the MLP projector and LoRA parameters.
Here $y_t$ denotes the token at position $t$ in the target frontier coordinate string (\eg ``\texttt{0.5, 3.2}''), and inference terminates upon generation of the coordinate delimiter.
The complete inference procedure is summarized in Algorithm~\ref{alg:main}.

\begin{algorithm}[t]
\caption{\ours Inference Pipeline (Per Episode)}
\label{alg:main}
\begin{algorithmic}[1]
\REQUIRE Navigation instruction $\mathcal{T}$, RGB-D stream $\{(I_t, D_t)\}$
\ENSURE Navigation trajectory $\tau$
\STATE Initialize visual memory buffer $\mathcal{M} \leftarrow \varnothing$
\FOR{each navigation step $t$}
  \STATE $V_t \leftarrow \text{Compress}(\text{DINOv3}(I_t))$ \hfill \COMMENT{Visual encoding}
  \STATE $\mathcal{M}_t \leftarrow \text{RollingBuffer}(\mathcal{M}, V_t, K)$ \hfill \COMMENT{Update memory}
  \STATE $\{q_i\}_{i=1}^{N}, \mathcal{F}_t \leftarrow \text{PQ3D}(I_t, D_t, p_t)$ \hfill \COMMENT{3D perception}
  \STATE $\{\hat{q}_i\} \leftarrow \text{MLPProjector}(\{q_i\})$ \hfill \COMMENT{Soft-token projection}
  \STATE $\mathcal{H} \leftarrow [\mathcal{M}_t;\; \hat{q}_1, \ldots, \hat{q}_{N};\; \text{Tokenize}(\mathcal{T})]$
  \STATE $\hat{c} \leftarrow \text{VLM.generate}(\mathcal{H})$ \hfill \COMMENT{VLM decision}
  \IF{$\mathcal{F}_t = \varnothing$}
    \STATE $f^{*} \leftarrow \text{Rescue}(p_t)$ \hfill \COMMENT{Fallback}
  \ELSE
    \STATE $f^{*} \leftarrow \text{SnapToFrontier}(\hat{c}, \mathcal{F}_t, r)$
  \ENDIF
  \STATE Execute $\text{NavigateTo}(f^{*})$
\ENDFOR
\end{algorithmic}
\end{algorithm}

\section{EXPERIMENTS}
\label{sec:experiments}

\subsection{Experimental Setup}
\label{sec:setup}

\subsubsection{Benchmarks}
We evaluate on the Habitat simulator~\cite{savva2019habitat} using three embodied navigation benchmarks that span a range of goal-directed tasks.
HM3D-OVON~\cite{yokoyama2024ovon} tests open-vocabulary object-goal navigation with three validation splits: \textit{val\_seen}, \textit{val\_seen\_synonyms}, and \textit{val\_unseen}, each containing 120 episodes (the official evaluation set defined by~\cite{yokoyama2024ovon}).
GOAT-Bench~\cite{khanna2024goat} evaluates multi-modal lifelong navigation in which each episode chains multiple goals (category, description, or image); following the evaluation protocol of~\cite{mtu3d}, we evaluate on 30 episodes per split.
SG3D~\cite{sg3d} tests scene-graph-grounded sequential navigation (338 episodes, 1{,}351 sub-tasks).

\subsubsection{Evaluation Protocol}
We report Success Rate ($\text{SR} = \frac{1}{N_{\text{total}}}\sum_{i=1}^{N_{\text{total}}} S_i$, where $S_i{=}1$ if the agent stops within $d_{\text{succ}}{=}0.25$\,m of the goal) and Success weighted by Path Length ($\text{SPL} = \frac{1}{N_{\text{total}}}\sum_{i=1}^{N_{\text{total}}} S_i \cdot \tfrac{l_i}{\max(p_i, l_i)}$, where $l_i$ is the shortest-path distance and $p_i$ the agent's actual path length).
For SG3D, SR is reported at sub-task (s-SR) and full-task (t-SR) granularities~\cite{sg3d}.

\subsubsection{Implementation Details}
All experiments use Habitat-Sim~\cite{savva2019habitat} with the Stretch embodiment~\cite{kemp2022stretch} (1.41\,m height, 0.17\,m base radius) following~\cite{yokoyama2024ovon,khanna2024goat}. The agent processes $360{\times}640$ RGB-D frames with discrete actions (forward 0.25\,m, turn left/right, look up/down) and a maximum of 500 steps per episode.
The MLP projector (${\sim}4$M parameters) uses learning rate $1{\times}10^{-4}$. LoRA~\cite{hu2022lora} adapters ($r{=}8$, $\alpha{=}32$, ${\sim}13$M parameters) are applied to \textit{q\_proj}, \textit{v\_proj}, \textit{down\_proj}, \textit{up\_proj} with learning rate $2{\times}10^{-5}$.
We train for 6 epochs on 1{,}187 SFT samples with effective batch size 16 (batch 4 $\times$ accumulation 4), using AdamW with linear warmup (100 steps) and cosine decay on a single NVIDIA RTX 4090.
PQ3D uses benchmark-specific stage-2 weights from MTU3D~\cite{mtu3d}; the navigation policy (MLP projector + LoRA) is trained once on OVON and shared across all benchmarks.

\subsubsection{Baselines}
We compare against prior embodied navigation methods spanning imitation learning, reinforcement learning, VLM-based, and 3D scene understanding paradigms~\cite{ramrakhya2023pirlnav, yokoyama2024vlfm, zhang2024uninavid, ziliotto2024tango, mtu3d}.
As an internal baseline, we include \textbf{Text-Hint}: the same PQ3D backend and VLM, but with 3D information transmitted as text-coordinate strings rather than soft tokens, using the base VLM without LoRA or additional training.
To isolate model adaptation from the information pathway, we also evaluate a Text-Hint + LoRA variant trained identically to \ours but receiving text coordinates instead of soft tokens (Table~\ref{tab:bottleneck}).
A \textbf{Rich-Text} variant further augments each object's text with distance, direction, and confidence descriptors to test whether enriching the text channel can narrow the representation gap (\S\ref{sec:ablation}).
Cross-paradigm comparisons in Table~\ref{tab:ovon} involve different training regimes; we therefore anchor causal claims on the MTU3D comparison, which shares the same frozen PQ3D encoder and benchmark-specific weights, and on the controlled ablation in Table~\ref{tab:bottleneck}, which additionally holds VLM and training data constant to isolate the transfer modality.

\subsection{Comparison with Prior Methods}
\label{sec:main_results}

\begin{table}[t]
\centering
\caption{Comparison on HM3D-OVON~\cite{yokoyama2024ovon}. SR (\%) and SPL (\%) are reported for 120 episodes per split with 0.25\,m success threshold. Best results in \textbf{bold}. ``--'' indicates unreported values.}
\label{tab:ovon}
\setlength{\tabcolsep}{3pt}
\resizebox{\columnwidth}{!}{%
\begin{tabular}{l cc cc cc}
\toprule
\multirow{2}{*}{Method} & \multicolumn{2}{c}{Val Seen} & \multicolumn{2}{c}{Val Synonyms} & \multicolumn{2}{c}{Val Unseen} \\
\cmidrule(lr){2-3}\cmidrule(lr){4-5}\cmidrule(lr){6-7}
 & SR & SPL & SR & SPL & SR & SPL \\
\midrule
BC~\cite{ramrakhya2023pirlnav} & 11.1 & 4.5 & 9.9 & 3.8 & 5.4 & 1.9 \\
DAgger~\cite{ramrakhya2023pirlnav} & 18.1 & 9.4 & 15.0 & 7.4 & 10.2 & 4.7 \\
BCRL~\cite{ramrakhya2023pirlnav} & 20.2 & 8.2 & 15.2 & 5.3 & 8.0 & 2.8 \\
VLFM~\cite{yokoyama2024vlfm} & 35.2 & 18.6 & 32.4 & 17.3 & 35.2 & 19.6 \\
DAgRL+OD~\cite{ramrakhya2023pirlnav} & 38.5 & 21.1 & 39.0 & 21.4 & 37.1 & 19.9 \\
RL~\cite{ramrakhya2023pirlnav} & 39.2 & 18.7 & 27.8 & 11.7 & 18.6 & 7.5 \\
DAgRL~\cite{ramrakhya2023pirlnav} & 41.3 & 21.2 & 29.4 & 14.4 & 18.3 & 7.9 \\
Uni-NaVid~\cite{zhang2024uninavid} & 41.3 & 21.1 & 43.9 & 21.8 & 39.5 & 19.8 \\
TANGO~\cite{ziliotto2024tango} & -- & -- & -- & -- & 35.5 & 19.5 \\
MTU3D~\cite{mtu3d} & 55.0 & 23.6 & 45.0 & 14.7 & 40.8 & 12.1 \\
\midrule
\textbf{\ours (Ours)} & \textbf{74.2} & \textbf{33.9} & \textbf{68.3} & \textbf{28.6} & \textbf{66.7} & \textbf{25.7} \\
\bottomrule
\end{tabular}%
}
\end{table}

\subsubsection{OVON Results}
Table~\ref{tab:ovon} compares \ours against prior methods on HM3D-OVON.
\ours achieves 74.2\% SR on \textit{val\_seen}, 68.3\% on \textit{val\_seen\_synonyms}, and 66.7\% on \textit{val\_unseen}, surpassing the strongest prior result (55.0\%/45.0\%/40.8\%) by +19.2\,pp, +23.3\,pp, and +25.9\,pp respectively.
Prior methods lack either structured 3D perception or VLM reasoning or both~\cite{ramrakhya2023pirlnav,yokoyama2024vlfm,zhang2024uninavid,ziliotto2024tango,mtu3d}; \ours is the first to combine both via soft-token injection.
\ours also attains the highest SPL across all splits (33.9\%/28.6\%/25.7\%), indicating that preserving continuous 3D representations improves path efficiency in addition to success rate.
The largest gains appear on \textit{val\_unseen} (+25.9\,pp SR), where the VLM's pretrained knowledge enables robust performance despite the limited fine-tuning set (${\sim}1{,}200$ samples), even on novel layouts that specialized policies struggle with.
Text-interface methods (SayNav~\cite{saynav}, SG-Nav~\cite{sg_nav}, L3MVN~\cite{l3mvn}) target fixed-vocabulary ObjectNav (6--21 categories) and are not directly comparable. AstraNav~\cite{astranav} uses a 1.0\,m success threshold and is likewise excluded.
\S\ref{sec:ablation} isolates the contribution of soft-token injection from model adaptation.

\subsubsection{Zero-Shot Generalization}
Tables~\ref{tab:goat} and~\ref{tab:sg3d} evaluate zero-shot policy transfer: the navigation policy (projector and adapters) is trained exclusively on OVON and applied without modification to GOAT-Bench and SG3D, while PQ3D uses benchmark-specific stage-2 weights following~\cite{mtu3d}.
On GOAT-Bench, \ours achieves 67.2\% SR on \textit{val\_seen} and 64.2\% on \textit{val\_unseen}, surpassing all reported baselines despite receiving no GOAT-specific training;
the consistency across all three splits (30 episodes each) provides corroborating evidence for the result.
On SG3D, \ours reaches 47.2\% s-SR (+23.4\,pp over the previous best) and 16.6\% t-SR (+8.6\,pp), indicating that soft-token injection generalizes across both multi-modal goals and sequential task structures.
On both benchmarks, SPL is slightly lower than the strongest baseline (GOAT: 29.0\% vs.\ 30.5\%; SG3D: 14.7\% vs.\ 16.5\%), reflecting a modest SR--SPL trade-off: the zero-shot policy achieves substantially higher success rates but follows somewhat longer paths in novel layouts, a pattern consistent with the per-episode analysis in \S\ref{sec:ablation}.
The controlled ablation confirms that SPL narrows from +9.2\,pp (\textit{val\_seen}) to $-2.6$\,pp (\textit{val\_unseen}) due to longer paths rather than degraded spatial reasoning (\S\ref{sec:ablation}).
These results suggest that PQ3D's query embeddings encode task-general navigation semantics that transfer across benchmark boundaries without retraining.

\begin{table}[t]
\centering
\caption{Zero-shot generalization to GOAT-Bench~\cite{khanna2024goat}. The navigation policy is trained only on OVON and transfers without modification. Baseline results are from~\cite{mtu3d,khanna2024goat,huang2025msgnav}. ``--'' indicates unreported values.}
\label{tab:goat}
\setlength{\tabcolsep}{3pt}
\resizebox{\columnwidth}{!}{%
\begin{tabular}{l cc cc cc}
\toprule
\multirow{2}{*}{Method} & \multicolumn{2}{c}{Val Seen} & \multicolumn{2}{c}{Val Synonyms} & \multicolumn{2}{c}{Val Unseen} \\
\cmidrule(lr){2-3}\cmidrule(lr){4-5}\cmidrule(lr){6-7}
 & SR & SPL & SR & SPL & SR & SPL \\
\midrule
Mod.\ CoW~\cite{gadre2023cow} & 14.8 & 8.7 & 18.5 & 11.5 & 16.1 & 10.4 \\
SA-NN Mono.~\cite{khanna2024goat} & 16.8 & 9.4 & 18.5 & 10.1 & 12.3 & 6.8 \\
Mod.\ GOAT~\cite{khanna2024goat} & 26.3 & 17.5 & 33.8 & 24.4 & 24.9 & 17.2 \\
SA-NN Chain~\cite{khanna2024goat} & 29.2 & 12.8 & 38.2 & 15.2 & 29.5 & 11.3 \\
TANGO~\cite{ziliotto2024tango} & -- & -- & -- & -- & 32.1 & 16.5 \\
MTU3D~\cite{mtu3d} & 52.2 & \textbf{30.5} & 48.4 & \textbf{30.3} & 47.2 & 27.7 \\
MSGNav~\cite{huang2025msgnav} & -- & -- & -- & -- & 52.0 & \textbf{29.6} \\
\midrule
\textbf{\ours (zero-shot)} & \textbf{67.2} & 29.0 & \textbf{63.2} & 25.8 & \textbf{64.2} & 24.7 \\
\bottomrule
\end{tabular}%
}
\end{table}

\begin{table}[t]
\centering
\caption{Zero-shot generalization to SG3D~\cite{sg3d}. Same zero-shot policy as GOAT-Bench. Baseline results are from~\cite{mtu3d,sg3d}.}
\label{tab:sg3d}
\begin{tabular}{l ccc}
\toprule
Method & s-SR & t-SR & SPL \\
\midrule
SenseAct-NN Mono.~\cite{sg3d} & 12.1 & 7.7 & 10.1 \\
Embodied VideoAgent~\cite{fan2024videoagent} & 14.7 & 3.8 & 10.2 \\
MTU3D~\cite{mtu3d} & 23.8 & 8.0 & \textbf{16.5} \\
\midrule
\textbf{\ours (zero-shot)} & \textbf{47.2} & \textbf{16.6} & 14.7 \\
\bottomrule
\end{tabular}
\end{table}

\subsection{Ablation Studies}
\label{sec:ablation}

\begin{table}[t]
\centering
\caption{Representation-gap ablation across all OVON splits. All rows share the same PQ3D backend and VLM. Soft tokens vs.\ Text-Hint: $\Delta$SPL $p{<}0.0001$, $\Delta$SR $p{=}0.028$ (details in Table~\ref{tab:significance}).}
\label{tab:bottleneck}
\setlength{\tabcolsep}{2.5pt}
\small
\begin{tabular}{@{}l cc cc cc@{}}
\toprule
\multirow{2}{*}{Input to VLM} & \multicolumn{2}{c}{Val Seen} & \multicolumn{2}{c}{Val Synonyms} & \multicolumn{2}{c}{Val Unseen} \\
\cmidrule(lr){2-3}\cmidrule(lr){4-5}\cmidrule(lr){6-7}
 & SR & SPL & SR & SPL & SR & SPL \\
\midrule
Text-Hint & 66.7 & 24.7 & 66.7 & 27.0 & 63.3 & 28.3 \\
Rich-Text (dist.+dir.+conf.) & 65.0 & 24.9 & 62.5 & 26.1 & 70.0 & 27.9 \\
Text-Hint + LoRA & 68.3 & 26.9 & 60.8 & 23.6 & 64.2 & 23.1 \\
\textbf{Soft tokens (\ours)} & \textbf{74.2} & \textbf{33.9} & \textbf{68.3} & \textbf{28.6} & \textbf{66.7} & \textbf{25.7} \\
\bottomrule
\end{tabular}
\end{table}

\subsubsection{Representation Gap}
Table~\ref{tab:bottleneck} isolates the information pathway: the 3D encoder and VLM are held constant while varying only the interface between them.
On \textit{val\_seen}, soft-token injection improves SPL from 24.7\% to 33.9\% (+9.2\,pp, $p{<}0.0001$) and SR from 66.7\% to 74.2\% (+7.5\,pp, $p{=}0.028$); statistical details appear in Table~\ref{tab:significance}.
Adding LoRA to the Text-Hint baseline yields only modest gains (+2.2\,pp SPL, +1.6\,pp SR), confirming that model adaptation alone cannot bridge the representation gap. The majority of the improvement stems from preserving continuous representations rather than from additional capacity.
PQ3D was chosen because its unified object-frontier query space directly supports soft-token injection; since Table~\ref{tab:bottleneck} holds the encoder constant, the relative advantage over text serialization is independent of encoder quality, though validating with alternative encoders remains important future work (\S\ref{sec:conclusion}).

Cross-split analysis reinforces this conclusion: on \textit{val\_seen\_synonyms}, LoRA adaptation with text coordinates \emph{decreases} SR by 5.9\,pp (from 66.7\% to 60.8\%), suggesting that fine-tuning on impoverished text representations amplifies overfitting rather than learning transferable strategies.
To test whether the gap can be narrowed by enriching the text channel, we evaluate a Rich-Text variant that augments each object with distance, egocentric direction, and confidence descriptors (Table~\ref{tab:bottleneck}).
On \textit{val\_seen}, Rich-Text achieves 65.0\% SR, statistically indistinguishable from the coordinate-only baseline (66.7\%, $p{=}0.78$) and significantly below \ours (74.2\%, $p{=}0.017$), indicating that the representation gap on familiar environments is modality-driven rather than information-driven.
On \textit{val\_unseen}, Rich-Text reaches 70.0\% SR, nominally above \ours (66.7\%) but statistically indistinguishable ($p{=}0.84$); across splits, Rich-Text shows inconsistent effects ($-1.7$/$-4.2$/$+6.7$\,pp SR), suggesting that text-conveyed spatial information is utilized unpredictably by the VLM.

\begin{figure}[t]
  \centering
  \begin{subfigure}[b]{0.48\columnwidth}
    \centering
    \includegraphics[width=\linewidth]{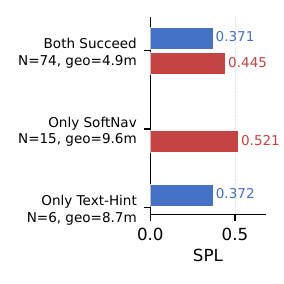}
    \caption{\textit{val\_seen}}
  \end{subfigure}
  \hfill
  \begin{subfigure}[b]{0.48\columnwidth}
    \centering
    \includegraphics[width=\linewidth]{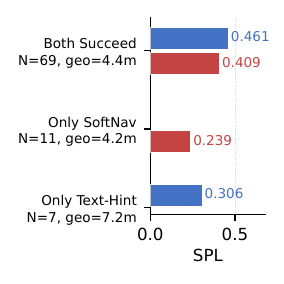}
    \caption{\textit{val\_unseen}}
  \end{subfigure}
  \caption{Per-episode SPL on OVON. Red: \ours; blue: Text-Hint. (a)~\textit{val\_seen}: \ours outperforms on shared successes ($p{=}0.039$) and solves 15 additional episodes at higher geodesic difficulty. (b)~\textit{val\_unseen}: \ours solves 11 exclusive episodes but with less direct paths on shared successes.}
  \label{fig:per_episode}
\end{figure}

\begin{figure}[t]
  \centering
  \begin{subfigure}[b]{0.31\columnwidth}
    \centering
    \includegraphics[width=\linewidth]{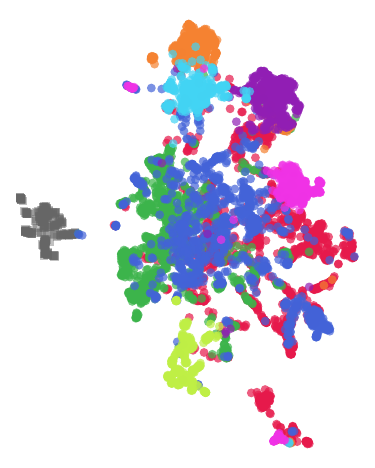}
    \caption{Raw 768-d}
  \end{subfigure}
  \hfill
  \begin{subfigure}[b]{0.31\columnwidth}
    \centering
    \includegraphics[width=\linewidth]{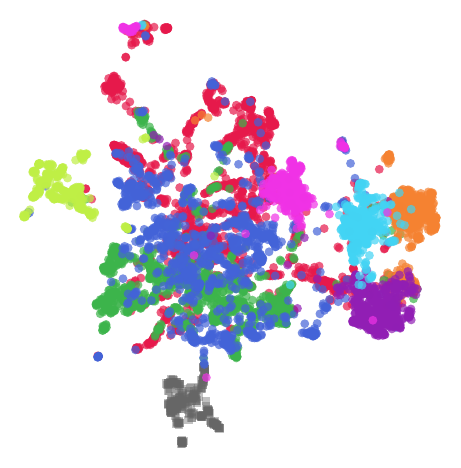}
    \caption{Projected 2048-d}
  \end{subfigure}
  \hfill
  \begin{subfigure}[b]{0.31\columnwidth}
    \centering
    \includegraphics[width=\linewidth]{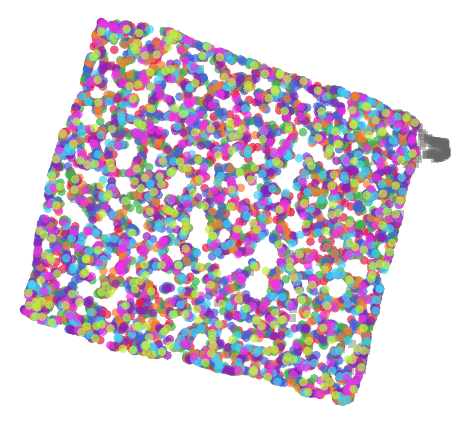}
    \caption{Text-serialized}
  \end{subfigure}
  \caption{UMAP visualization of PQ3D query embeddings (50 episodes). Colors denote K-means clusters. (a)~Raw 768-d embeddings form clear semantic clusters. (b)~MLP projection to 2048-d preserves cluster structure. (c)~Text serialization destroys all organization.}
  \label{fig:tsne}
\end{figure}

Per-episode analysis (Fig.~\ref{fig:per_episode}) reveals split-dependent behavior. On \textit{val\_seen}, \ours achieves higher SPL on 74 shared-success episodes (0.445 vs.\ 0.371, $p{=}0.039$) and additionally solves 15 episodes with nearly double the geodesic distance (9.6\,m vs.\ 4.9\,m, $p{<}0.001$)---a dual advantage in both path efficiency and coverage. On \textit{val\_unseen}, the 11 newly solved episodes have comparable geodesic distance to shared successes (4.2\,m vs.\ 4.4\,m, $p{=}0.61$), and the SPL gap ($-2.6$\,pp) arises primarily from longer paths on 69 shared-success episodes (0.409 vs.\ 0.461). This contrast suggests that while soft tokens provide semantic cues benefiting success rate across splits, leveraging their spatial cues for optimal path planning relies heavily on LoRA adaptation quality. Expanding the training set could further improve spatial adaptation in novel environments.
Fig.~\ref{fig:tsne} provides visual confirmation: PQ3D embeddings exhibit clear semantic clustering that the MLP projector preserves, while text serialization destroys all structure.

\begin{figure*}[t]
  \centering
  \includegraphics[width=\textwidth]{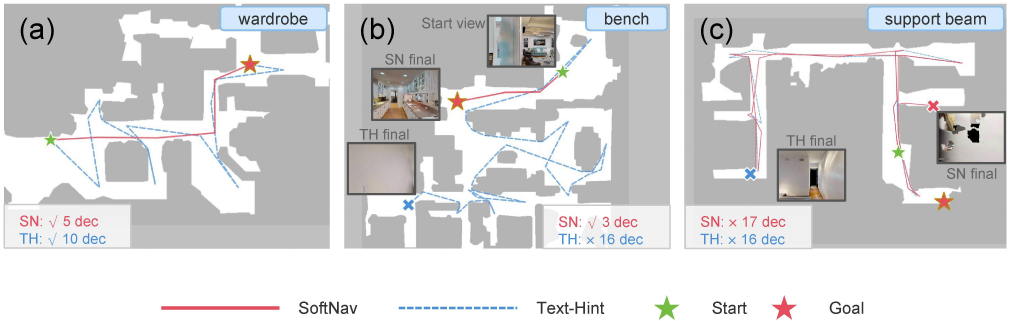}
  \caption{Representative trajectories on OVON \textit{val\_unseen}. (a)~Both succeed; \ours is more efficient. (b)~Only \ours succeeds. (c)~Both fail---the support beam is heavily occluded, limiting 3D encoder cues.}
  \label{fig:qualitative}
\end{figure*}

\begin{figure*}[t]
  \centering
  \includegraphics[width=\textwidth]{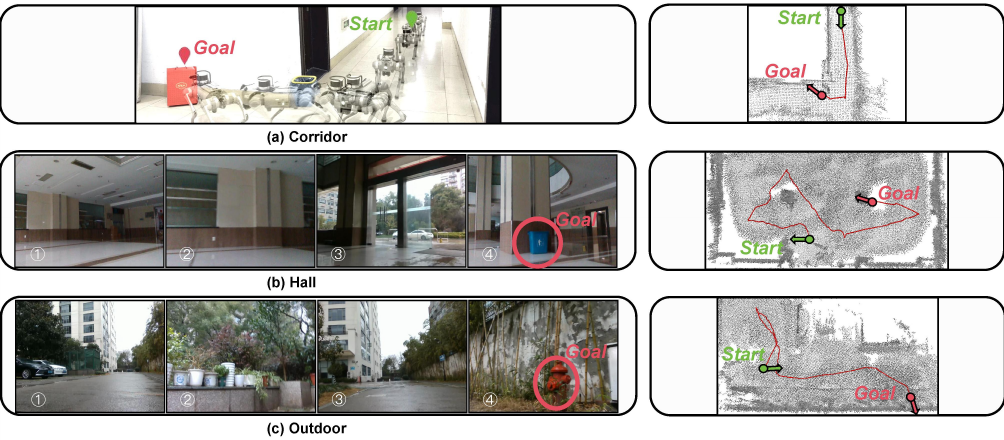}
  \caption{Real-world deployment on a Unitree Go2. (a)~Corridor (8/10): multi-exposure overlay and point cloud map. (b)~Hall (6/10) and (c)~Outdoor (5/10): onboard observations and point cloud maps with trajectories. The outdoor environment lies outside the indoor training distribution.}
  \label{fig:realworld}
\end{figure*}

\begin{table}[t]
\centering
\caption{Token selection ablation on OVON \textit{val\_seen}. $K$: number of tokens retained after selection. The full model ($K{=}128$, no selection) performs best.}
\label{tab:selection}
\begin{tabular}{@{}lccc@{}}
\toprule
Variant & $K$ & SR & SPL \\
\midrule
Full model & 128 & \textbf{74.2} & \textbf{33.9} \\
w/ selector ($K{=}32$) & 32 & 73.3 & 32.3 \\
w/ selector ($K{=}96$) & 96 & 68.3 & 30.6 \\
\bottomrule
\end{tabular}
\end{table}

\subsubsection{Token Selection}
Table~\ref{tab:selection} explores an Instruction-Conditioned Token Selector that uses cross-attention between the navigation instruction and PQ3D query tokens to select the top-$K$ most relevant tokens before projection.
Neither $K{=}32$ (retaining 25\% of tokens) nor $K{=}96$ (75\%) improves over injecting all 128 tokens.
This suggests that, under our training regime, PQ3D query tokens are densely informative for navigation; aggressive selection discards useful information.
An alternative explanation is that the hard top-$K$ operation limits gradient flow to the selector, and the training set (1{,}187 samples) may be insufficient to simultaneously optimize the selector, projector, and LoRA adapters.

\begin{table}[t]
\centering
\caption{Statistical significance of \ours vs.\ Text-Hint on OVON \textit{val\_seen}. Paired bootstrap with $N{=}10{,}000$ resamples.}
\label{tab:significance}
\begin{tabular}{@{}lcccc@{}}
\toprule
Metric & $\Delta$ & 95\% CI & $p$-value \\
\midrule
$\Delta$SPL & +0.092 & $[0.041, 0.144]$ & $<$0.0001 \\
$\Delta$SR & +0.075 & $[0.000, 0.150]$ & 0.028 \\
\midrule
\multicolumn{4}{l}{\small \ours SR 95\% CI: $[0.658, 0.817]$; lower bound exceeds} \\
\multicolumn{4}{l}{\small the strongest prior result (0.550) by +10.8\,pp.} \\
\bottomrule
\end{tabular}
\end{table}

\subsubsection{Statistical Significance}
Table~\ref{tab:significance} reports paired bootstrap results ($N{=}10{,}000$) for the core ablation.
The SPL confidence interval lies entirely above zero; notably, even the lower bound of the 95\% CI for \ours SR (65.8\%) exceeds the strongest prior result (55.0\%) by +10.8\,pp, indicating that the performance advantage is robust under worst-case sampling variability.

\subsection{Analysis}
\label{sec:analysis}

\subsubsection{Inference Efficiency}
Each decision step takes ${\sim}1$\,s on a single RTX 4090: PQ3D 3D perception (${\sim}0.45$\,s) and VLM inference (${\sim}0.49$\,s) account for nearly all computation, while the MLP projector adds only 0.07\,ms.
The VLM cost is inherent to all methods that employ language-model reasoning for navigation~\cite{zhou2024navgpt,navid2024}; the soft-token injection mechanism itself introduces no measurable overhead.
The projector architecture is compatible with any VLM backbone or 3D encoder that produces fixed-dimensional embeddings, enabling substitution as more efficient architectures become available.

\subsubsection{Qualitative Analysis}
Fig.~\ref{fig:qualitative} presents three representative episodes from \textit{val\_unseen}.
When both methods succeed~(a), \ours follows a more direct path (5 vs.\ 10 decisions), suggesting that continuous embeddings support more efficient frontier selection.
In~(b), \ours reaches the target in 3 decisions, while the text-serialized baseline explores away from the goal across 16 decisions and exhausts the step budget.
In the failure case~(c), both methods explore extensively without locating the target, illustrating a shared limitation: when the 3D encoder cannot reliably detect the goal object, neither transfer pathway compensates.

\subsubsection{Real-World Testing}
To evaluate sim-to-real transfer, we deploy \ours on a Unitree Go2 quadruped robot equipped with an Intel RealSense D435i RGB-D camera.
The navigation policy trained entirely in Habitat is applied without domain adaptation or fine-tuning; perception and planning are processed offboard, and the Go2's built-in locomotion controller executes waypoint navigation to each selected frontier.
We test 10 episodes in each of three environments: an indoor corridor, a building hall, and an outdoor area; success is judged by a human operator when the robot stops near the target object.
The agent succeeds in 19 of 30 trials (63.3\%): 8/10 in the corridor, 6/10 in the hall, and 5/10 outdoors---notably, the outdoor environment lies entirely outside the indoor training distribution yet the agent still achieves a modest success rate.
These results confirm that the soft-token injection pipeline transfers to physical deployment without architectural modification.

\section{CONCLUSION}
\label{sec:conclusion}

We have identified and quantified a representation gap in VLM-based navigation: text serialization discards geometric and semantic content that direct embedding-level transfer preserves.
\ours bridges this gap by projecting entity-level 3D representations into the VLM's hidden space via a lightweight MLP projector, requiring only ${\sim}17$M trainable parameters and ${\sim}1{,}200$ samples.
The resulting policy achieves state-of-the-art SR and SPL on HM3D-OVON, and generalizes zero-shot to GOAT-Bench, SG3D, and real-world robot deployment (including outdoor environments beyond the training distribution) without retraining.
The controlled ablation confirms that neither enriching the text channel nor model adaptation alone closes this gap, indicating that these representations encode task-general navigation semantics.
Future work includes validating with alternative 3D encoders, scaling the training set, and adopting more efficient architectures to reduce the current ${\sim}1$\,s-per-step latency.

\bibliographystyle{IEEEtran}

\end{document}